\documentclass[conference]{IEEEtran}
\IEEEoverridecommandlockouts
\usepackage{cite}
\usepackage{amsmath,amssymb,amsfonts}
\usepackage{algorithm}
\usepackage{algorithmic}
\usepackage{graphicx}
\usepackage{textcomp}
\usepackage{xcolor}
\usepackage[utf8]{inputenc}
\usepackage[greek,english]{babel}
\usepackage{alphabeta}
\usepackage{soul}
\usepackage{subcaption}
\usepackage[export]{adjustbox}
\usepackage{makecell}
\usepackage{xcolor}
\definecolor{cream}{RGB}{255,253,240}
\usepackage{booktabs}
\usepackage{tikz}
\usetikzlibrary{arrows.meta,positioning,fit,calc,shapes.geometric}

\renewcommand{\baselinestretch}{0.978}

\newcommand\freefootnote[1]{%
  \let\thefootnote\relax%
  \footnotetext{#1}%
  \let\thefootnote\svthefootnote%
}

\usepackage{fancyhdr}
 
\def\BibTeX{{\rm B\kern-.05em{\sc i\kern-.025em b}\kern-.08em
    T\kern-.1667em\lower.7ex\hbox{E}\kern-.125emX}}
\begin{document}

\title{A Physics-Guided Transformer Framework for Electromigration Analysis in Multi-Segment Interconnects}

\author{
\IEEEauthorblockN{Pavlos Stoikos}
\IEEEauthorblockA{Dept. of Electrical and Computer Engineering\\
University of Thessaly, Greece\\
Email: pastoikos@e-ce.uth.gr}
\and
\IEEEauthorblockN{Anuj Pathania}
\IEEEauthorblockA{University of Amsterdam\\
Amsterdam, The Netherlands\\
Email: a.pathania@uva.nl}
\and
\IEEEauthorblockN{George Floros}
\IEEEauthorblockA{Dept. of Electronic and Electrical Engineering,\\
 Trinity College Dublin, Ireland \\
Email: florosg@tcd.ie}
}

\maketitle


\begin{abstract}
As technology scales to smaller nodes, increasing current densities make electromigration (EM) one of the dominant reliability challenges in on-chip interconnects.
Accurate transient stress analysis is needed to identify wires susceptible to EM degradation, but applying physics-based solvers across many interconnects remains computationally expensive.
This paper proposes a physics-guided transformer framework for fast EM stress prediction in multi-segment interconnect lines.
The framework converts each line into geometry- and DC-aware segment tokens and uses transformer attention to capture line-level context. A lightweight query decoder then predicts stress at selected locations and time instants.
The model is trained with an objective that combines normalized supervised regression, linewise relative-$L_2$ loss, and physics-guided continuity and terminal-flux terms.
Experiments on IBM power grid benchmarks show that the proposed model achieves relative-$L_2$ error below 8\% and reaches up to 2459.68$\times$ speedup compared with the matrix exponential~solver.
\end{abstract}


\section{Introduction}

Electromigration (EM) has become one of the greatest concerns for the semiconductor industry in recent years.  EM failures 
constitute an inevitable consequence
of the rising current demands and the smaller process geometries, and may lead to a number of open- or short-circuits in on-chip interconnects \cite{yang_2017}.
Classical criteria such as Black's equation and the Blech criterion~\cite{Black1969,Blech1976} provide useful lifetime guidelines, while physics-based solvers based on Korhonen-type stress diffusion models provide more accurate transient stress solutions~\cite{Korhonen1993,cook2018,chen-semi-analytical,10044796}. However, repeatedly solving the transient problem over many interconnect lines and reliability time instants remains computationally expensive.

\freefootnote{*This work was funded under the COIN-3D project, which has received funding from the European Union's Horizon Europe research and innovation program under grant agreement No. 101159667.}

Learning-based EM predictors address this cost by learning the mapping from interconnect features to stress response. Graph-based models and hierarchical physics-informed neural networks have shown promising speedups~\cite{Jin2021EMGraph,Chen2022HierPINN}. However, EM stress prediction remains challenging because the stress response is influenced by both local segment properties and the broader context of the interconnect~line.

In this work, we propose a physics-guided transformer framework for fast EM stress prediction in multi-segment interconnect lines. The main contributions are: (i) a transformer-based predictor for variable-length line sequences; (ii) DC-aware segment representations and a query-based decoder for stress prediction at multiple time instants; (iii) a physics-guided objective with continuity and terminal-flux regularization; and (iv) evaluation on IBM power grid benchmarks, demonstrating low stress-prediction error and large speedups over the matrix EM exponential solver~\cite{10044796}.

\begin{figure*}[!t]
    \centering
    \includegraphics[width=\textwidth]{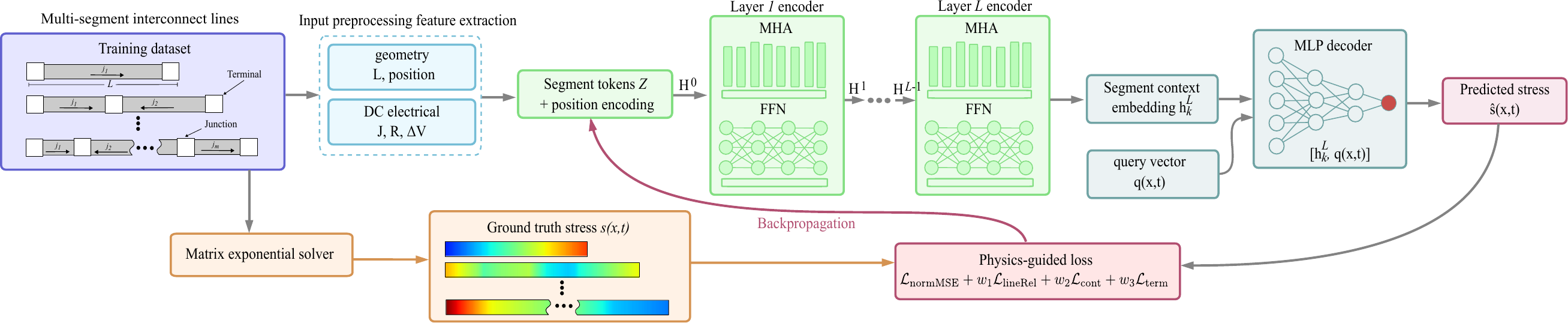}
    \caption{Architecture and training flow of the proposed physics-guided transformer framework.}
    \label{fig:physics_em_framework}
\end{figure*}
\section{Background}

EM stress evolution in confined metal interconnects is commonly described through Korhonen's stress diffusion model~\cite{Korhonen1993}. In one-dimensional form, the hydrostatic stress $\sigma(x,t)$ evolves according to
\[
\frac{\partial \sigma}{\partial t}
=
\frac{\partial}{\partial x}
\left[
\kappa
\left(
\frac{\partial \sigma}{\partial x}
+ \beta j
\right)
\right], 
\]
where $\sigma$ is the hydrostatic stress, $x$ is the position along the segment, $t$ is time, and $j$ is
the segment current density. The coefficients are given by $\beta=e\rho Z^{*}/\Omega$ and $\kappa=D_a\mathcal{B}\Omega/(k_BT)$, with $D_a=D_0\exp(-E_a/k_BT)$ denoting the atomic diffusion coefficient. Here, $e$ is the elementary charge, $\rho$ is the metal resistivity, $Z^{*}$ is the effective charge number, $\Omega$ is the atomic volume, $\mathcal{B}$ is the bulk modulus,
Th$k_B$ is Boltzmann's constant, $T$ is the absolute temperature, $D_0$ is the pre-exponential diffusivity constant, and $E_a$ is the activation energy.
e term $\beta j$ represents the EM
driving contribution, while $\partial\sigma/\partial x$ represents the opposing
back-stress contribution.
Their balance determines the net atomic flux and the resulting stress buildup along the interconnect.

In Cu dual-damascene (Cu DD) technology, diffusion barriers confine atomic transport within individual metallization layers, effectively suppressing mass exchange across adjacent layers~\cite{5510581}. As a result, large interconnect networks such as power grids can be analyzed layer by layer. Within each layer, the routing topology can be decomposed into multi-segment interconnect lines composed of wire segments and vias~\cite{6549125}. Each segment carries a locally defined current density, making segment-level physical and electrical features central to EM stress prediction.

Within such multi-segment interconnect lines, the stress at a queried point
depends not only on local segment properties, including geometry and electrical
quantities, but also on the surrounding segment sequence and terminal behavior.
Consequently, effective EM prediction requires models that capture both local
segment-level features and broader line-level context.
\section{Physics-guided transformer framework}

In this section, we present the proposed physics-guided transformer framework for fast EM stress prediction in multi-segment interconnect lines.
The overall flow is shown in Fig.~\ref{fig:physics_em_framework}.
Our methodology consists of four main stages: (i) each extracted interconnect line 
is represented as a sequence of segment-level physical and electrical features,
(ii) reference stress labels are generated at selected space-time query points,
(iii) a transformer encoder learns contextual representations of all segments
in the line, and (iv) a query-based decoder predicts the EM stress under a
combined supervised and physics-guided training objective.

\subsection{Feature construction and reference labels}


Each interconnect line is represented as an ordered sequence of
$K$ segments.
This representation preserves the physical ordering of the line and naturally supports different numbers of segments.
For segment $k$, we construct an input token $\mathbf{z}_k$ from normalized
geometry, line position, and DC electrical quantities,
\[
\mathbf{z}_k =
\left[
\ell_k,\,
p_k,\,
I_k,\,
\tilde{I}_k,\,
R_k,\,
\Delta V_k,\,
R_{\mathrm{line}},\,
\Delta V_{\mathrm{line}}
\right],
\]
where $\ell_k$ is the normalized segment length, $p_k$ is the relative segment
position within the line, $I_k$ is the signed segment current, $\tilde{I}_k$
is the clipped and normalized current, $R_k$ and $\Delta V_k$ are the segment
resistance and voltage drop, and $R_{\mathrm{line}}$ and
$\Delta V_{\mathrm{line}}$ are the total line resistance and voltage drop.
These DC-derived features provide the model with the dominant current-flow
information that drives EM stress buildup.

Reference stress labels are generated using the matrix exponential EM
solver~\cite{10044796}. For each sampled location $x$ along a segment and
aging time $t$, the solver returns the reference stress label $\sigma(x,t)$.
This label-generation procedure supports $N$ discrete reliability time
instants sampled from the lifetime interval $[0,20]$ years.

\subsection{Transformer architecture for EM stress prediction}

The segment tokens are first projected to a hidden dimension and combined with positional encodings to form the input sequence to the transformer encoder.
Each encoder block consists of multi-head self-attention (MHA) followed by a position-wise feed-forward network (FFN), with residual connections and layer normalization applied around both sublayers.
Self-attention allows each segment to aggregate information from all other segments in the same line, which is important because EM stress at a point depends not only on the local current but also on the surrounding segment configuration and terminal behavior.
After $L$ encoder layers, the model produces contextual segment embeddings
\[
    \mathbf{H}^{L}=\{\mathbf{h}_1^{L},\ldots,\mathbf{h}_K^{L}\},
\]
where $\mathbf{h}_k^{L}$ summarizes both the local features of segment $k$ and the line-level context.

Stress prediction is carried out by a local query decoder.
For a query point on segment $k$, the decoder combines the contextual embedding
$\mathbf{h}_k^{L}$ with a query vector $\mathbf{q}(x,t)$. This vector encodes
the normalized local coordinate within the segment and the selected aging time.
The concatenated representation is then passed through a two-hidden-layer MLP
with GELU activations and a final linear layer to produce the scalar
stress prediction,
\[
\hat{\sigma}(x,t) = D_\theta\left([\mathbf{h}_k^L,\mathbf{q}(x,t)]\right),
\]
where $D_\theta$ denotes the learnable decoder.

This design separates line-level context extraction from pointwise stress evaluation.
The transformer learns how segments interact along the line, while the decoder evaluates stress at arbitrary sampled locations and time instants.

\subsection{Physics-guided objective}

The predictor is trained with a compound objective that combines normalized supervised regression with line-level and boundary-aware physical regularization:
\[
\mathcal{L}
=
\mathcal{L}_{\mathrm{normMSE}}
+
w_1 \mathcal{L}_{\mathrm{lineRel}}
+
w_2 \mathcal{L}_{\mathrm{cont}}
+
w_3 \mathcal{L}_{\mathrm{term}} .
\]
Let $i$ index a line, $q$ index a valid query point on that line, and $Q_i$ denote the number of valid queries.
The reference stress is denoted by $\sigma_{iq}$ and the predicted stress by $\hat{\sigma}_{iq}$.
To improve training stability across benchmarks with different stress magnitudes, the supervised regression term is computed on normalized stresses,
$
\tilde{\sigma}_{iq}=\frac{\sigma_{iq}-\mu_{\sigma}}{s_{\sigma}},
\qquad
\hat{\tilde{\sigma}}_{iq}=\frac{\hat{\sigma}_{iq}-\mu_{\sigma}}{s_{\sigma}},
$
where $\mu_\sigma$ and $s_\sigma$ denote the training-set stress offset and scale, respectively.
The normalized MSE term is calculated as:
\[
\mathcal{L}_{\mathrm{normMSE}}
=
\frac{1}{\sum_i Q_i}
\sum_i \sum_{q=1}^{Q_i}
\left(\hat{\tilde{\sigma}}_{iq}-\tilde{\sigma}_{iq}\right)^2 .
\]

In addition to pointwise fitting, we include a linewise relative-$L_2$ term so that full stress profiles are optimized as line-level objects:
\[
\mathcal{L}_{\mathrm{lineRel}}
=
\frac{1}{|\mathcal{B}|}
\sum_{i\in\mathcal{B}}
\frac{
\sum_{q=1}^{Q_i}\left(\hat{\sigma}_{iq}-\sigma_{iq}\right)^2
}{
\max\left(
\sum_{q=1}^{Q_i}\sigma_{iq}^{2},
Q_i\sigma_{\mathrm{floor}}^{2}
\right)
},
\]
where $\mathcal{B}$ is the minibatch and $\sigma_{\mathrm{floor}}$ prevents near-zero-stress lines from dominating the objective.
The continuity term penalizes stress mismatch at internal segment interfaces,
\[
\mathcal{L}_{\mathrm{cont}}
=
\frac{1}{|\mathcal{E}|}
\sum_{(i,k)\in\mathcal{E}}
\left(
\hat{\tilde{\sigma}}_{i,k}(1,t)
-
\hat{\tilde{\sigma}}_{i,k+1}(0,t)
\right)^2 ,
\]
where $\mathcal{E}$ denotes the set of internal interfaces between adjacent
segments, $k$ indexes the segment position within line $i$, and $x=1$ and
$x=0$ correspond to the right endpoint of segment $k$ and the left endpoint
of segment $k+1$, respectively.
Finally, the terminal-flux term regularizes the predicted stress slope at the two blocking terminals,
\[
\mathcal{L}_{\mathrm{term}}
=
\frac{1}{|\mathcal{T}|}
\sum_{(i,e)\in\mathcal{T}}
\left(
\frac{\partial \hat{\tilde{\sigma}}_{i,e}}{\partial x}
-
\frac{-\beta J_{i,e} L_{i,e}}{s_{\sigma}}
\right)^2 ,
\]
where $\mathcal{T}$ contains the left and right terminal segments, $e$ indexes
a terminal segment of line $i$, and $J_{i,e}$ and $L_{i,e}$ denote its current
density and length.
In implementation, the derivative is approximated with a short finite difference near each endpoint.

\subsection{Hyperparameter search}


We use Optuna~\cite{akiba2019optuna} for hyperparameter selection, 
with validation relative-$L_2$ error as the search objective. For an evaluation
set $\mathcal{D}$, this metric is defined as:
\[
\mathrm{rel}\text{-}L_2(\mathcal{D})
=
100
\sqrt{
\frac{
\sum_{(i,q)\in\mathcal{D}}
\left(\hat{\sigma}_{iq}-\sigma_{iq}\right)^2
}{
\sum_{(i,q)\in\mathcal{D}}
\sigma_{iq}^2
}}
.
\]
The tuned parameters include the transformer width,
number of encoder layers, decoder hidden dimension, learning rate,
batch size, and the physics-loss weights.
Promising trials are trained for up to 600 epochs with
periodic validation and pruning of unpromising configurations. After selection,
the best configuration is evaluated on benchmark lines that were not used for
supervised training.

\section{Evaluation}

\subsection{Experimental setup}

We evaluate the proposed model on interconnect lines extracted from IBM power grid benchmarks.
For each line, stress labels are generated with the reference solver~\cite{10044796}.
The evaluation is performed at three reliability time instants,
$t\in\{5,10,20\}$ years, and the reported accuracy metrics are averaged over these time points.
All training, hyperparameter optimization, and inference experiments were run on a Linux PyTorch server equipped with an NVIDIA RTX 6000 Ada Generation GPU.
Across the IBM benchmarks, Optuna~\cite{akiba2019optuna} selected encoder hidden dimensions
$\{96,128,160\}$, 2-4 MHA/FFN layers, decoder hidden dimensions $\{96,128,192\}$, batch sizes $\{8,16\}$, and learning rates in
$[8.7{\times}10^{-5},\,9.6{\times}10^{-4}]$.

\begin{table}[!htb]
\centering
\caption{IBM line-dataset statistics.}
\vspace{-0.6em}
\label{tab:ibm_dataset_stats}
\resizebox{\columnwidth}{!}{%
\begin{tabular}{cccccc}
\toprule
Benchmark & Total lines & Train/all (\%) & Max seg. & Mean seg. & Max current (Pa)\\
\midrule
IBM1 & 4160 & 30 & 55 & 7 & $1.15{\times}10^{9}$ \\
IBM2 & 1202 & 60 & 236 & 104 & $9.14{\times}10^{8}$ \\
IBM3 & 17513 & 23 & 965 & 47 & $6.40{\times}10^{8}$ \\
IBM4 & 26302 & 15 & 572 & 35 & $1.58{\times}10^{7}$ \\
IBM5 & 6483 & 62 & 577 & 166 & $1.03{\times}10^{8}$ \\
IBM6 & 23595 & 41 & 1010 & 69 & $4.06{\times}10^{8}$ \\
\bottomrule
\end{tabular}}
\vspace{-0.6em}
\end{table}

For every benchmark, the generated stress labels and the train/test split are cached once and reused across all training runs.
This avoids rerunning the reference solver during model selection and makes the comparisons reproducible.
For supervised learning, we select a compact representative subset from each IBM benchmark.
The subset is chosen to represent the benchmark's line population while reducing the amount of reference data required for supervised training.
The remaining extracted lines are held out from supervised training and used only for evaluation.
Table~\ref{tab:ibm_dataset_stats} summarizes the extracted-line statistics for each benchmark, where the \textit{Train/all} column denotes the percentage of extracted lines used for supervised training.

\subsection{Representative-subset results}

\begin{table}[!b]
\centering
\caption{Evaluation results for the IBM line datasets summarized in Table~\ref{tab:ibm_dataset_stats}.}
\vspace{-0.6em}
\label{tab:representative_subset_results}
\resizebox{\columnwidth}{!}{%
\begin{tabular}{ccccc}
\toprule
Benchmark & Train rel-$L_2$ (\%) & Test rel-$L_2$ (\%) & Test RMSE (Pa) & Test MAE (Pa) \\
\midrule
IBM1 & 7.3 & 6.8 & $1.97{\times}10^{3}$ & $1.20{\times}10^{3}$ \\
IBM2 & 6.1 & 7.0 & $7.15{\times}10^{2}$ & $2.38{\times}10^{2}$ \\
IBM3 & 5.8 & 7.8 & $2.72{\times}10^{5}$ & $6.73{\times}10^{4}$ \\
IBM4 & 5.2 & 5.7 & $6.29{\times}10^{2}$ & $2.00{\times}10^{2}$ \\
IBM5 & 4.4 & \textbf{3.7} & $3.18{\times}10^{1}$ & $1.86{\times}10^{1}$ \\
IBM6 & 3.8 & \textbf{4.2} & $1.90{\times}10^{4}$ & $7.31{\times}10^{3}$ \\
\bottomrule
\end{tabular}}
\vspace{-0.8em}
\end{table}

We report relative-$L_2$ error, root-mean-square error (RMSE), and mean absolute error (MAE).
Relative-$L_2$ measures the normalized stress-profile error, while RMSE and MAE report absolute stress deviations in pascals.
Table~\ref{tab:representative_subset_results} summarizes the accuracy of the proposed model on the representative IBM subsets.
Across all six benchmarks, the test relative-$L_2$ remains below 8\%, showing that the selected training lines capture the dominant stress behavior of the extracted line population.
The train and test relative-$L_2$ values are also close for most benchmarks, which indicates that the proposed model does not simply memorize the supervised subset.
The strongest relative accuracy is obtained on IBM5 and IBM6, while IBM2 and IBM3 remain the most challenging cases due to their larger structural variation and wider stress-scale range.
This is also reflected in the absolute metrics, where IBM3 exhibits larger RMSE and MAE even though its relative error remains comparable to the other benchmarks.



\begin{figure}[!htb]
\centering
\includegraphics[width=\columnwidth]{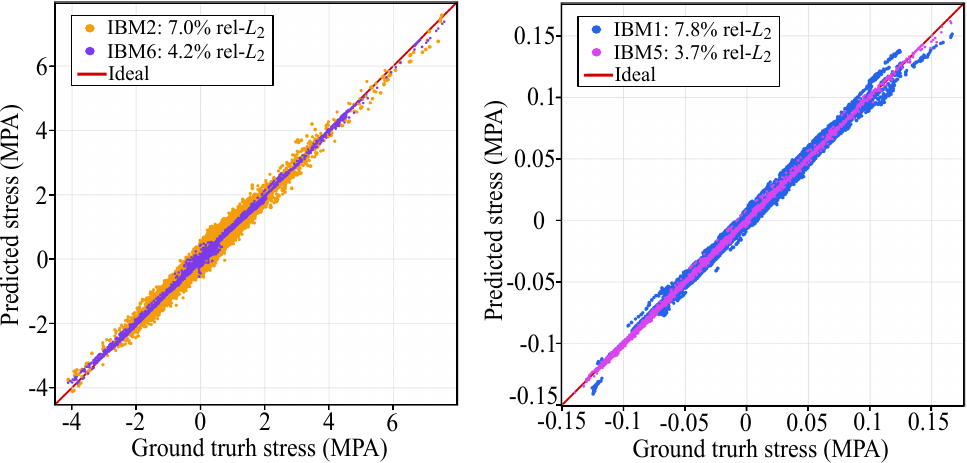}
\vspace{-0.8em}
\caption{Predicted-versus-ground-truth stress comparison on representative IBM test cases.}
\vspace{-0.8em}
\label{fig:pred_truth_scatter}
\end{figure}


\begin{figure}[!htb]
\centering
\includegraphics[width=\columnwidth]{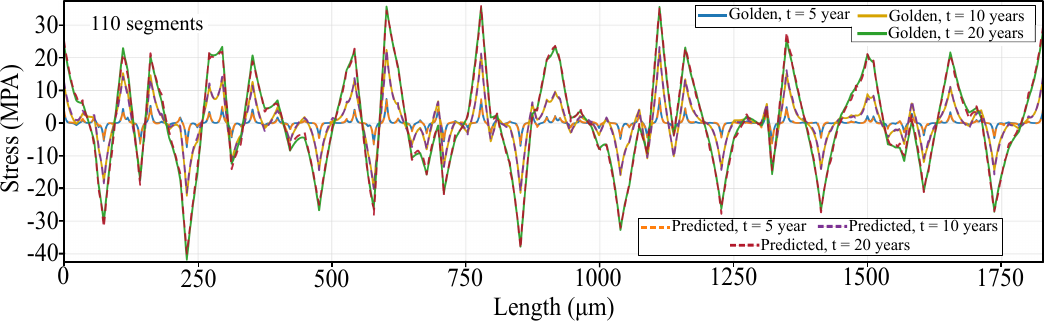}
\vspace{-0.8em}
\caption{Stress evolution along a 110-segment interconnect for multiple time points.}
\vspace{-0.8em}
\label{fig:stress_profile_110seg}
\end{figure}

The prediction quality is illustrated in Fig.~\ref{fig:pred_truth_scatter}  by comparing the model
outputs with the reference stresses. The close alignment of the
scatter points with the ideal diagonal shows that the proposed model
preserves the stress scale across different IBM benchmarks, despite their
different line lengths and stress ranges. A complementary profile-level
view is shown in Fig.~\ref{fig:stress_profile_110seg} for a representative 110-segment interconnect,
where the predicted curves track the reference solution across multiple
time points and reproduce the dominant stress peaks, including localized
variations near segment transitions.

\subsection{Runtime comparison}

\begin{table}[!b]
\centering
\caption{Average runtime comparison over the 10 largest extracted lines of each IBM benchmark.}
\vspace{-0.6em}
\label{tab:runtime_comparison}
\scriptsize
\resizebox{\columnwidth}{!}{%
\begin{tabular}{cccc}   
\toprule
Benchmark & Golden (s/line) & Proposed (s/line) & Speedup \\
\midrule
IBM1 & 0.019 & $5.21{\times}10^{-5}$ & 367.95x \\
IBM2 & 0.139 & $5.65{\times}10^{-5}$ & \textbf{2459.68x} \\
IBM3 & 0.350 & $8.61{\times}10^{-4}$ & 406.24x \\
IBM4 & 0.222 & $3.11{\times}10^{-4}$ & 712.86x \\
IBM5 & 0.224 & $1.33{\times}10^{-4}$ & 1675.76x \\
IBM6 & 0.389 & $5.51{\times}10^{-4}$ & 706.53x \\
\bottomrule
\end{tabular}%
}
\vspace{-0.6em}
\end{table}

Table~\ref{tab:runtime_comparison} compares the inference runtime of the trained model and the reference solver.
To approximate a worst-case runtime setting, both methods are evaluated on the 10 largest extracted lines of each IBM
benchmark, and the table reports the average runtime over these cases.
For the proposed method, the measured time corresponds to a forward pass of the trained model, while the reference time corresponds to computing the reference stress solution for the same line.
Training and hyperparameter search are offline one-time costs and are therefore not included in this per-line comparison.
This setting reflects the intended deployment use case, where the trained model is reused to evaluate many new interconnect lines.
The largest acceleration is observed on IBM2, with up to 2459.68$\times$ speedup.
Even in the least favorable case, IBM1, the proposed model remains more than two orders of magnitude faster than the reference solver.

\section{Conclusion}

This paper presented a physics-guided transformer framework for fast EM stress prediction in multi-segment interconnect lines.
The model encodes each line as geometric and DC-aware segment tokens and uses a query decoder to predict stress at selected locations and time instants.
To improve the physical consistency of the predicted stress profiles, the
training objective includes continuity and terminal-flux terms.
Experiments on IBM power-grid benchmarks show that the model achieves test
relative-$L_2$ error below 8\%, while providing
an average speedup of 1054.84$\times$ over the matrix exponential solver.
Future work will extend the framework to interconnect-tree topologies and
investigate adaptive selection of samples for label generation.

\begingroup
\scriptsize
\bibliographystyle{IEEEtran}
\bibliography{bibliography}
\endgroup


\end{document}